\documentclass[twoside,lettersize,journal]{IEEEtran}

\usepackage{amsmath,amsfonts}
\usepackage[linesnumbered, ruled,vlined]{algorithm2e}
\usepackage{algorithmic}
\algsetup{linenosize=\tiny}
\usepackage{url}
\usepackage{verbatim}
\usepackage{graphicx}
\usepackage{makecell}
\usepackage{multirow}
\usepackage{wrapfig}
\usepackage{hyperref}
\usepackage{microtype}
\usepackage{tablefootnote}

\usepackage{amssymb}
\usepackage{pifont}
\newcommand{\cmark}{\ding{51}}%
\newcommand{\xmark}{\ding{55}}%

\makeatletter
\newcommand{\disablehyperlinks}{\hypersetup{hidelinks, draft}}
\newcommand{\enablehyperlinks}{\hypersetup{hidelinks=false, draft=false}}
\makeatother

\usepackage{tikz,pgfplots}
\usetikzlibrary{positioning}
\usetikzlibrary{backgrounds}
\usetikzlibrary{matrix}
\usepgfplotslibrary{groupplots}
\pgfplotsset{compat=newest}
\usepgfplotslibrary{fillbetween}

\usepackage{lineno}

\newcommand{\DNNShifter}{\texttt{DNNShifter }}

\newcommand{\DNNShifterr}{\texttt{DNNShifter}}

\newcommand{\bailey}[1]{\textcolor{black}{#1}}

\usepackage{xcolor}
\usepackage[tikz]{bclogo}
\usepackage[framemethod=tikz]{mdframed}
\usepackage{lipsum}
\usepackage[many]{tcolorbox}

\definecolor{bgblue}{RGB}{245,243,253}
\definecolor{ttblue}{RGB}{91,194,224}

\mdfdefinestyle{mystyle}{%
  rightline=true,
  innerleftmargin=10,
  innerrightmargin=10,
  outerlinewidth=3pt,
  topline=false,
  rightline=true,
  bottomline=false,
  skipabove=\topsep,
  skipbelow=\topsep
}

\newtcolorbox{myboxii}[1][]{
  breakable,
  freelance,
  title=#1,
  colback=white,
  colbacktitle=white,
  coltitle=black,
  fonttitle=\bfseries,
  bottomrule=0pt,
  boxrule=0pt,
  colframe=white,
  overlay unbroken and first={
  \draw[black!75!black,line width=1.5pt]
    ([xshift=5pt]frame.north west) -- 
    (frame.north west) -- 
    (frame.south west);
  \draw[black!75!black,line width=1.5pt]
    ([xshift=-5pt]frame.north east) -- 
    (frame.north east) -- 
    (frame.south east);
  },
  overlay unbroken app={
  \draw[black!75!black,line width=1.5pt,line cap=rect]
    (frame.south west) -- 
    ([xshift=5pt]frame.south west);
  \draw[black!75!black,line width=1.5pt,line cap=rect]
    (frame.south east) -- 
    ([xshift=-5pt]frame.south east);
  },
  overlay middle and last={
  \draw[black!75!black,line width=1.5pt]
    (frame.north west) -- 
    (frame.south west);
  \draw[black!75!black,line width=1.5pt]
    (frame.north east) -- 
    (frame.south east);
  },
  overlay last app={
  \draw[black!75!black,line width=1.5pt,line cap=rect]
    (frame.south west) --
    ([xshift=5pt]frame.south west);
  \draw[black!75!black,line width=1.5pt,line cap=rect]
    (frame.south east) --
    ([xshift=-5pt]frame.south east);
  },
}

\begin{document}

\title{\DNNShifterr: An Efficient DNN Pruning System for Edge Computing}
\author{Bailey~J.~Eccles,
        Philip~Rodgers,
        Peter~Kilpatrick,
        Ivor~Spence,
        and~Blesson~Varghese%
\thanks{B. Eccles and B. Varghese are with the School of Computer Science, University of St Andrews, UK. (e-mail: bje1@st-andrews.ac.uk)}%
\thanks{P. Kilpatrick and I. Spence are with the School of Electronics, Electrical Engineering and Computer Science, Queen’s University Belfast, UK.}%
\thanks{P. Rodgers is with Autonomous Networking Research \& Innovation Department, Rakuten Mobile, Japan.}%
}

\maketitle

\begin{abstract}
Deep neural networks (DNNs) underpin many machine learning applications. Production quality DNN models achieve high inference accuracy by training millions of DNN parameters which has a significant resource footprint. This presents a challenge for resources operating at the extreme edge of the network, such as mobile and embedded devices that have limited computational and memory resources. To address this, models are pruned to create lightweight, more suitable variants for these devices. \bailey{Existing pruning methods are unable to provide similar quality models compared to their unpruned counterparts without significant time costs and overheads or are limited to offline use cases. Our work rapidly derives suitable model variants while maintaining the accuracy of the original model. The model variants can be swapped quickly when system and network conditions change to match workload demand.} This paper presents \DNNShifterr, an end-to-end DNN training, spatial pruning, and model switching system that addresses the challenges mentioned above. At the heart of \DNNShifter is a novel methodology that prunes sparse models using structured pruning \bailey{- combining the accuracy-preserving benefits of unstructured pruning with runtime performance improvements of structured pruning}. The pruned model variants generated by \DNNShifter are \bailey{smaller in size and thus faster than dense and sparse model predecessors, making them suitable for inference at the edge while retaining near similar accuracy as of the original dense model.} \DNNShifter generates a portfolio of model variants that can be swiftly interchanged depending on operational conditions. \DNNShifter produces pruned model variants up to 93x faster than conventional training methods. Compared to sparse models, the pruned model variants are up to 5.14x smaller and have a 1.67x inference latency speedup, with no compromise to sparse model accuracy. 
In addition, \DNNShifter has \bailey{up to 11.9x lower} overhead for switching models \bailey{and up to 3.8x} lower memory utilisation than existing approaches. \bailey{\DNNShifter is available for public use from \url{https://github.com/blessonvar/DNNShifter}}.
\end{abstract}








\section{Introduction}
\label{sec:introduction}
Deep neural networks (DNNs) are machine learning (ML) models comprising a sequence of layers, such as convolution and linear. Such models find application in object detection and image classification due to their high accuracy~\cite{b1}. Production quality DNN models trained on standard datasets contain a large number of parameters. For example, VGG-16~\cite{b3} trained on the ImageNet~\cite{b2} dataset contains 138M parameters. Such models have significant CPU or memory resource requirements and, consequently, high energy consumption for training and inference. Hence, they are suited for resource-rich environments like cloud or high-performance computing sites. These DNNs cannot be adopted for relatively resource-constrained environments, such as the (extreme) network edge dominated by mobile and embedded devices~\cite{b4}.

Edge environments cannot support production quality DNNs due to compute~\cite{b4}, memory~\cite{b8} and energy~\cite{edc} constraints.
Therefore, approaches for deriving lightweight DNN model variants from production quality DNNs using (i) neural architecture search (NAS)~\cite{b14} and (ii) pre-trained model compression~\cite{b8} have been proposed. These approaches have a two-fold \textbf{\textit{limitation}}. Firstly, they are time-consuming and costly~\cite{b14}. For example, the NasNet~\cite{nasnet} search requires four days of computation on 500 data centre-grade GPUs to find optimal model variants.

Secondly, the model variants obtained from these approaches are static. The models are optimised against specific objectives, such as accuracy, inference latency, or model size~\cite{b14}. Therefore, they cannot be used on the edge to meet the requirements of varying operational conditions, such as changing resource utilisation levels~\cite{model-switch, dynamic-ofa}.

Existing NAS and compression approaches cannot be used for rapidly producing models, and the models produced by these approaches cannot be adapted to suit changing operational conditions on the edge.
The research reported in this paper is therefore focused towards addressing the above limitations and surmounts the following challenges:

\textit{\textbf{Challenge 1} - Rapidly generating a range of DNNs suited for different operational conditions and heterogeneous edge resources:} ML applications that run on the edge will need to execute pre-trained DNNs. Training a DNN tailored to the edge resource using approaches, such as NAS, is not suitable as they are time and energy-consuming~\cite {b14}. Alternatively, compressing a pre-trained DNN using knowledge distillation~\cite{kd} is based on trial and error, or quantisation~\cite{quant} that requires specialised hardware or libraries that may not be available in edge environments.

\textit{\textbf{Challenge 2} - Spatial compression of DNN models while maintaining accuracy:} DNN compression methods, such as structured pruning~\cite{b13} or re-parameterisation~\cite{repvgg}, can significantly reduce the size of a model. However, these methods remove parameters essential to model accuracy. For example, convolutional layers are sensitive to pruning, and even small degrees of pruning can negatively impact accuracy~\cite{b13}. Consequently, the compressed model is fine-tuned after pruning using computationally expensive methods to regain accuracy, which can take up to 3 times the original training time~\cite{b8}.

\textit{\textbf{Challenge 3} - On-demand switching of compressed DNNs to adapt to dynamic operational environments:}
DNNs used at the edge will need to seamlessly adapt to changing conditions in real time by switching models on-demand to match model inference performance thresholds.
However, existing approaches will incur a downtime in the order of minutes~\cite{b14} to hours~\cite{easiedge} for identifying and deploying a suitable model that meets the desired performance~\cite{dynamic-ofa}.

This paper presents \DNNShifterr, a framework that utilises production quality sparse models to generate a portfolio of spatially compressed model variants \bailey{with high accuracy} in real time. This is achieved by proposing a novel method that uses structured pruning of highly sparse models. This results in pruned models with a smaller resource footprint and the same model accuracy as the original sparse model. This method fundamentally differs from the commonly reported structured pruning methods that prune pre-trained dense models and negatively impact accuracy. 
The portfolio of models that are generated by our method can be used to adapt to match a range of operational runtime requirements. This is achieved by low overhead switching from one model to another in the portfolio at runtime. 
\DNNShifter makes the following three research \textbf{\textit{contributions}}:

1) A time and resource-efficient guided DNN model-training, pruning, and runtime switching pipeline that creates a portfolio of spatially pruned model variants comparable to a typical DNN model training routine using NAS. \bailey{\DNNShifter generates a portfolio of pruned model variants up to 93x faster than state-of-the-art methods.}

2) A novel pruning method to compress highly sparse DNN models, resulting in accurate and spatially compact pruned model variants suited for edge resources with low inference latency. \bailey{\DNNShifter pruned model variants are up to 5.14x smaller and have up to 1.67x and 1.45x faster CPU and GPU inference latencies, respectively. In addition, the pruned model variants can be obtained orders of magnitude faster when compared to existing structured pruning methods and have higher accuracy for a given model size.}

3) A low-overhead method that switches from one model variant to another on-demand at the edge to match a range of operational runtime requirements. \bailey{\DNNShifter has up to 11.9x lower overhead for switching model variants with up to 3.8x lower memory utilisation than existing approaches.}


The remainder of this paper is organised as follows. Section~\ref{sec:background} discusses related work. Section~\ref{sec:system} presents the \DNNShifter framework. Section~\ref{sec:experimentalstudies} presents experimental results. \bailey{Section~\ref{sec:conclusions} concludes the paper by discussing system limiations.}

\section{Related work}
\label{sec:background}
Approaches for DNN compression aim to improve the resource efficiency of models by reducing their computational and memory utilisations while preserving the model's accuracy. Techniques such as model pruning, quantisation, and knowledge distillation leverage different properties of a DNN that inherently lend towards compression. \bailey{As a result, a compressed model that is either smaller, faster or both compared to the original model, is produced.} These approaches typically produce a single compressed model. 

Techniques such as NAS, on the other hand, generate a portfolio of compressed models from a search space that suits the requirements of constrained resources~\cite{nasnet, liu2018darts, prenas, fgcs2}. However, \bailey{NAS} is computationally expensive because it trains and evaluates many candidate models (up to thousands) before identifying the optimal set of compressed models. \bailey{\textit{Our work is positioned at the intersection of DNN compression and NAS, where a more time and resource-efficient compression pipeline than NAS is developed to generate a portfolio of highly compressed models, which serves a range of resource requirements as seen in edge environments}}. \bailey{This section considers the key contributions of each DNN compression method by providing an overview of their strengths and weaknesses and comparing their features. More recent work is considered to highlight the novelty of the \DNNShifter framework we propose in addition to presenting baseline methods to compare \DNNShifter in the experiments presented in Section~\ref{sec:experimentalstudies}.}

\subsection{Unstructured pruning and sparse models}
Unstructured pruning masks select individual parameters of the DNN by setting their weight to zero~\cite{obd, bg2, smf, ntksap}. \bailey{Existing methods} such as Lottery Ticket Hypothesis (LTH)~\cite{b16} demonstrate that introducing sparsity via unstructured pruning early into training can lead to similar or higher final accuracy as a dense model. In addition, with suitable hardware and sparse matrix libraries~\cite{cer}, sparse models can accelerate model training, thus, reducing time and energy costs~\cite{sparse_accel}. \bailey{The concept of LTH has motivated a large collection of unstructured pruning methods~\cite{bg2, ntksap, b17, b20}.} However, resource-constrained environments usually do not support libraries to leverage any performance benefits of sparse models~\cite{benefits}. Furthermore, the zeroed weights within the sparse model do not reduce the memory footprint but create irregular memory accesses that will degrade inference performance on conventional CPUs. \bailey{Unstructured pruning research typically focuses on improving sparse model accuracy rather than considering compute performance considerations~\cite{b17}}. \bailey{\textit{Sparse models are the starting point for our work}}. The \DNNShifter framework removes sparse data structures within these models in a disciplined manner. In other words, we spatially remove the zeroed weights, thereby reducing the model size and inference latency while maintaining the same accuracy as the original sparse model.

\begin{figure}[!t]
\centerline{\includegraphics[width=0.3\textwidth]{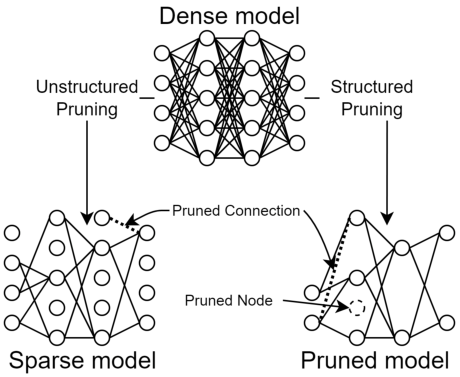}}
\caption{Obtaining sparse and pruned models from pruning a dense model.}
\label{pruning}
\end{figure}

\subsection{Structured pruning and re-parameterisation}
\bailey{As shown in Figure~\ref{pruning},} structured pruning spatially removes groups of parameters, such as convolutional filters~\cite{b13, easiedge, b12, prospr, fgcs1}. A ranking algorithm is employed to identify filters that contribute the least to accuracy loss if removed. Structured pruning removes these filters, resulting in models with lower memory, energy, and inference footprint. However, structured pruning is time-consuming because: (i) the filters that can be removed need to be identified given the thousands of pruning combinations~\cite{easiedge, b12}, and (ii) the parameters that remain after pruning are fine-tuned to recover the accuracy lost while pruning~\cite{ b8,bg2}. Thus, on-demand compression cannot be achieved using structured pruning. 

However, \DNNShifter achieves structured pruning in \bailey{(near)} real-time by leveraging the following observations of sparse models: (i) zeroed weights are inherently ranked as prunable, and (ii) pruning of zeroed weights does not reduce model accuracy. Furthermore, structured re-parameterisation can be combined with structured pruning to further optimise a model by modifying the underlying architecture for a target device. For example, RepVGG~\cite{repvgg} restructures ResNets into a VGG-like architecture and improves GPU utilisation.

\subsection{Dynamic DNNs}
Dynamic DNNs improve inference efficiency by adapting a DNN for different operational conditions~\cite{b22}. Methods such as skipping layers~\cite{skipnet} and connections~\cite{helios} or early-exiting~\cite{b22} decrease inference latency at the cost of inference accuracy. Although dynamic DNNs offer the advantage of using any sub-model from within a single model, there are no spatial benefits since the entire model runs in memory, even for a small sub-model~\cite{b22}. Alternatively, \DNNShifter provides both inference and spatial benefits and leverages in-memory compression of multiple sparse models to facilitate on-demand switching of models to suit runtime requirements.

\subsection{Other compression methods}
Other compression methods, namely quantisation and knowledge distillation, are presented in the literature for DNN compression. Quantisation reduces the bit precision of parameters in DNNs to reduce the model size and to accelerate inference~\cite{quant}. However, quantisation is usually applied to all parameters of the DNN, which incurs an accuracy loss. Furthermore, quantised models may require dedicated hardware to \bailey{carry out} inference at a lower precision. Knowledge distillation transfers \bailey{training} knowledge from a more extensive teacher to a smaller student model~\cite{kd}. The student model achieves similar accuracy to the teacher model and is spatially smaller. However, knowledge distillation is not easily automated to serve various model architectures and produces only a single student model rather than a portfolio of models \bailey{suited for different operational conditions, such as specific memory budgets.} Therefore, knowledge distillation does not scale for the varying resource requirements of deployments seen in heterogeneous edge environments.

\begin{table}[t]
\caption{\bailey{Comparison of Neural Architecture Search (NAS), Unstructured Pruning (UP), Structured Pruning (SP), Dynamic DNNs (D-D), and the proposed framework, \DNNShifterr.}
	}
	\label{tab:comp}
    \centering
    {\color{black}
    \begin{tabular}{p{3cm}ccccc}

\hline
                  & \footnotesize{NAS} & \footnotesize{UP} & \footnotesize{SP} & \footnotesize{D-D} & \footnotesize{\DNNShifter} \\ \hline
\footnotesize{Generates model variants to suit different operational conditions}    &  \cmark   &  \xmark  &  \xmark  &     \cmark      &     \cmark       \\
\footnotesize{Produces memory efficient models}  &  \xmark   &  \xmark  &  \cmark  &    \xmark      &     \cmark       \\
\footnotesize{Produces high accuracy models}     &  \cmark   & \cmark   & \xmark   &  \cmark        &   \cmark         \\
\footnotesize{Produces compute efficient models} &  \xmark   & \xmark   & \cmark   &   \xmark       &    \cmark        \\
\footnotesize{On-demand switching of models at runtime} &  \xmark   &  \xmark  &  \xmark  &  \cmark        &    \cmark        \\\hline

\end{tabular}
}
\end{table}   

\subsection{\color{black}{Addressing open gaps with our contribution}}
\bailey{Although existing compression methods have a range of benefits, they present one or more significant limitations that prohibit their use for on-demand deployment of production quality DNNs to edge devices. \textit{\DNNShifter leverages the accuracy-preserving benefits of unstructured pruning with the runtime performance improvements of structured pruning across various model sizes to suit different operational conditions seen in edge environments. This combination has not been previously explored in the literature}. \DNNShifter creates an efficient training, pruning, and inference pipeline, which is highlighted in comparison to other DNN compression methods in Table~\ref{tab:comp}. 
The \DNNShifter framework and the models generated by the framework meet the requirements for deploying DNNs in edge environments. However, existing methods have one or more limitations making them less suited for edge systems.  
The next section explores the underlying methodology and implementation of \DNNShifterr.}

\begin{figure*}[!t]
\centerline{\includegraphics[width=1.0\textwidth]{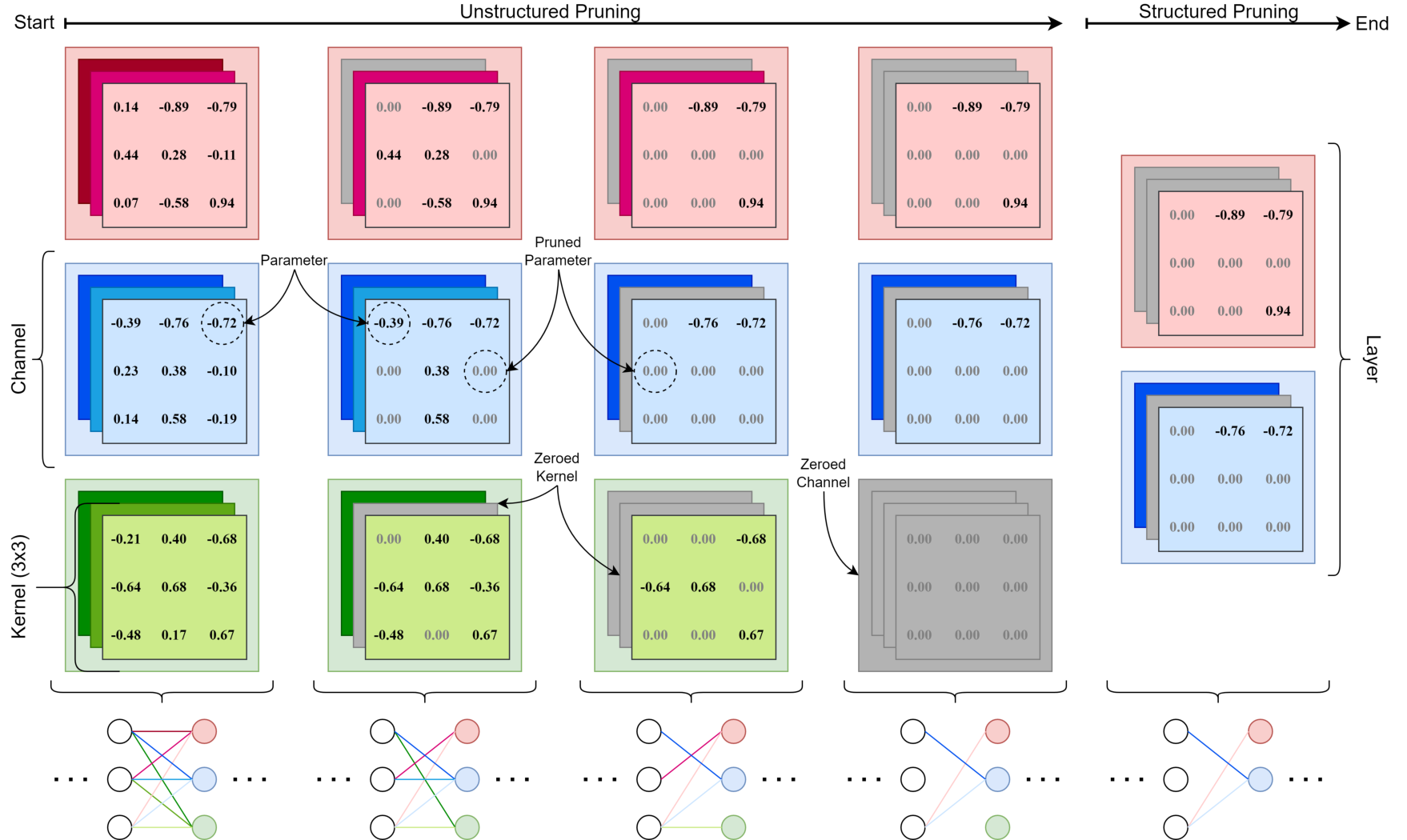}}
\caption{Structured pruning zero-valued data structures obtained from unstructured pruning.}
\label{ussp}
\end{figure*}

\section{DNNShifter}
\label{sec:system}
\DNNShifter is a framework that can be employed for resource-constrained environments, such as at the network edge or the extreme edge that has relatively limited computational capabilities. The framework prunes production quality DNN models on-demand and provides automated runtime model switching for inference on constrained resources. \DNNShifter can be employed by system administrators for managing the life cycle of ML application development, deployment, and simulation environments to address the following challenges:

\textbf{Rapidly obtaining production quality models:} In real-time, \DNNShifter offers structured pruning of large sparse DNN models that cannot be run on hardware-limited resources. The framework derives pruned model variants for the target resource without a significant accuracy loss while achieving this \bailey{on a small} monetary, computation, and energy \bailey{budget}. This contrasts existing approaches that employ NAS~\cite{b14} or parameter fine-tuning~\cite{b12}.

\textbf{Hardware agnostic automated model pruning:} \DNNShifter creates a portfolio of hardware-independent pruned model variants with different performance characteristics (e.g. model size, speed, and accuracy) to suit all deployment conditions. The model variants can be deployed across different resource tiers based on operational conditions, such as resource availability and performance targets. The approach adopted by \DNNShifter is hardware agnostic and is not specific to specialised hardware such as a GPU~\cite{sparse_accel}.

\textbf{Real-time model switching at runtime:} Once a model portfolio has been deployed on the target hardware, \DNNShifter utilises the portfolio of pruned model variants to \bailey{select a model variant suited for a given operational condition. The framework facilitates the adaptation of the model to suit variations in the operational conditions with low overheads.} The underlying method in \DNNShifter switches the active model for inference from the portfolio via inflation (which activates the pruned model) and deflation (which further compresses and deactivates the pruned model) to match operational demand. 

\DNNShifter is envisioned to be a holistic end-to-end solution for ML edge applications that reduces human administrator interventions or domain-specific knowledge for creating pruned models. \DNNShifter can also benchmark different pruning algorithms on heterogeneous hardware and make informed decisions in the life cycle of edge-based ML applications.

This section will further present the observations that motivated the design of \DNNShifter and provides an overview of the framework. 
\subsection{Motivation}\label{motivation}
A variety of model pruning methods have been presented in the literature for reducing the complexity of production models to suit resource-constrained environments while maintaining accuracy~\cite{bg2}. Traditional pruning methods are limited in multiple ways: (a) many require further time-consuming fine-tuning after the initial pruned model is obtained~\cite{b12}, (b) many rely on hardware accelerators~\cite{sparse_accel}, and (c) pruning often requires a costly trial and error process to determine the optimal pruned model for a given target resources~\cite{b22}. 
\bailey{Current pruning methods are unsuitable for real-time execution in critical scenarios, such as on-device video analytics, that require sub-second latency to preserve optimal service quality~\cite{realtime}}. 

\DNNShifter was developed to address the above limitations by leveraging the following two observations:

\subsubsection{Aggregating and pruning unstructured sparsity} 
\label{patterns}
During unstructured pruning, aggregating parameters from sparse data structures will result in fully prunable data structures that directly impact the model size and inference latency. Figure~\ref{ussp} highlights this observation. During unstructured pruning, the parameters of a convolutional kernel are set to zero values. The data structures (matrices) representing the kernels may be sparse (not all values are zero) and, therefore, cannot be pruned without compromising accuracy (shown as unprunable data structure). Zero matrices are obtained as pruning progresses by using the parameter ranking algorithm used in unstructured pruning. A data structure that has full zero values is prunable and, by employing structured pruning, can be removed from the model. This results in reducing the model size and, thereby, inference latency.

\bailey{\textit{\DNNShifter leverages this observation to prune sparse models using structural pruning without degrading model accuracy. Since model accuracy is preserved, \DNNShifter does not require fine-tuning after pruning. }}

\subsubsection{Further compression of remaining model sparsity} During runtime, inactive models from the portfolio can be further compressed to reduce overheads. After structured pruning, the remaining unprunable data structures contain sparse matrices (Figure~\ref{ussp}). Sparse matrices have repeating and compressible data patterns (of zeroed weights). Therefore, the model can be encoded (deflated) into smaller representations while the model is inactive. For example, such deflation may be applied when downloading the model portfolio from a cloud/edge server to target device hardware or, during runtime, to models in a portfolio that are not actively inferring.

\bailey{\textit{\DNNShifter uses this observation to load the entire portfolio of deflated models into memory during runtime. When a specific model is required for inference, it is decoded (inflated) in (near) real-time. This allows model switching in response to varying operational conditions on the edge and is significantly faster than existing methods.}}
\subsection{Framework overview} 
\label{overview}

\begin{figure*}[ht]
\centerline{\includegraphics[width=0.95\textwidth]{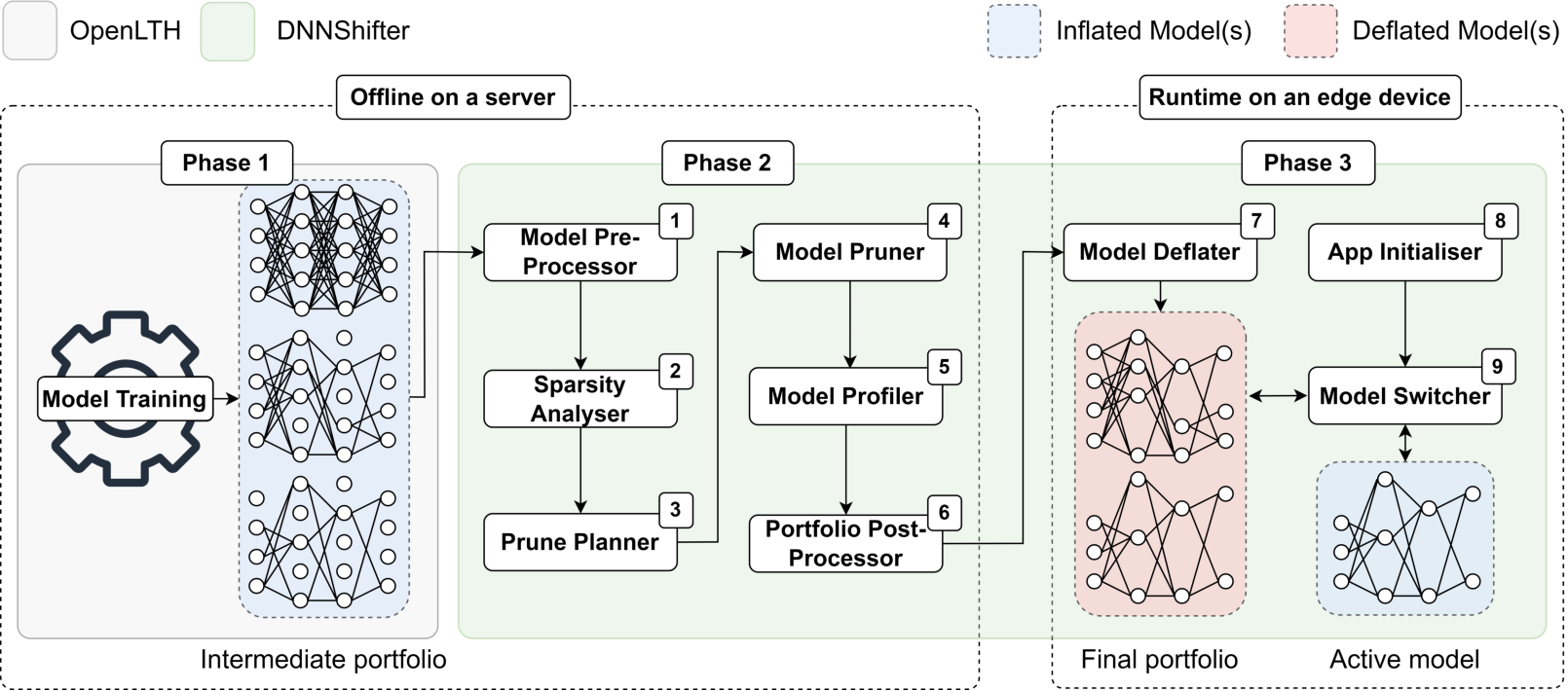}}
\caption{\bailey{Overview of the \DNNShifter framework.}}
\label{fig_conversion}
\end{figure*}

This section presents \bailey{an overview of the \DNNShifter framework. It operates in three phases, as shown in Figure~\ref{fig_conversion}}: 

\textit{Phase 1: Offline training of production quality DNNs with unstructured ranking} - In this phase, model training and parameter ranking are combined into a single iterative training process. A production-quality DNN model is taken as input by an unstructured pruning method. Then, the insignificant parameters of the model are masked between each training iteration by an unstructured ranking method to produce a portfolio of model variants (one per iteration) with sparse data structures (referred to as sparse models). 

\textit{Phase 2: On-demand conversion from sparse models to pruned models} - The portfolio of sparse models is pruned via structured pruning to obtain pruned model variants that can be deployed on a target hardware resource. 

\textit{Phase 3: Runtime model switching} - The portfolio is deployed, monitored, and adapts to varying operational conditions by switching the active pruned model variant at runtime. 

\bailey{Phase 1 builds on an existing technique and Phase 2 and Phase 3 comprise nine modules. The next sections discuss these.}
\subsection{Phase 1: Model training and parameter ranking} 
\label{phase1}
This phase uses an unstructured ranking algorithm to produce highly sparse models from production-quality DNNs (dense models) while maintaining viable model accuracy. It is to be noted that the sparse models obtained from this phase will include zero values in the model parameters. However, since they are not removed from the data structures until the next phase, sparse models are not smaller in size than the dense model. 
Choosing an unstructured ranking algorithm over structured pruning eliminates the need for fine-tuning after training to recover model accuracy~\cite{b13, b12}. In addition, such a ranking approach between training iterations has two advantages. 

Firstly, \DNNShifter simplifies model parameter ranking \bailey{so that a user does not require expert knowledge of ranking algorithms and no additional hyperparameters need be configured.} 

Secondly, \DNNShifter improves the model pruning pipeline efficiency. A conventional pruning pipeline consists of training the model, compressing using structured pruning methods, profiling, and iteratively fine-tuning the pruned model for the target hardware. While training and compression can occur offline on large-scale computational resources, fine-tuning will need to be carried out on the target hardware that may be resource-limited. 
The final accuracy of the model cannot be determined until this time-consuming pipeline is completed. If the desired accuracy is not obtained, then the entire sequence of the pruning pipeline must be repeated with a different set of pruning hyperparameters. In addition, only a single pruned model will be obtained at the end of the pipeline that meets specific operational conditions. If the operational condition changes, the entire pruning pipeline must be repeated. 

\DNNShifter improves the efficiency of the pruning pipeline in three ways by integrating ranking within the training iterations: (i) The final model accuracy that can be achieved is known before the pipeline completes. Therefore, the pipeline can be reinitialised in advance with new hyperparameters if the target accuracy cannot be achieved. (ii) 
Fine-tuning, a computationally intensive task, is eliminated on the target hardware resource that may be relatively resource constrained. Therefore, rapid and on-demand deployments of DNN models are feasible since fine-tuning does not need to be carried out. (iii) A portfolio of pruned models can be generated that will suit a range of operational conditions on the target hardware resource by running the pruning pipeline once. This allows for adapting a model that is deployed at runtime. 

\begin{figure}[!t]
\centerline{\includegraphics[width=0.45\textwidth]{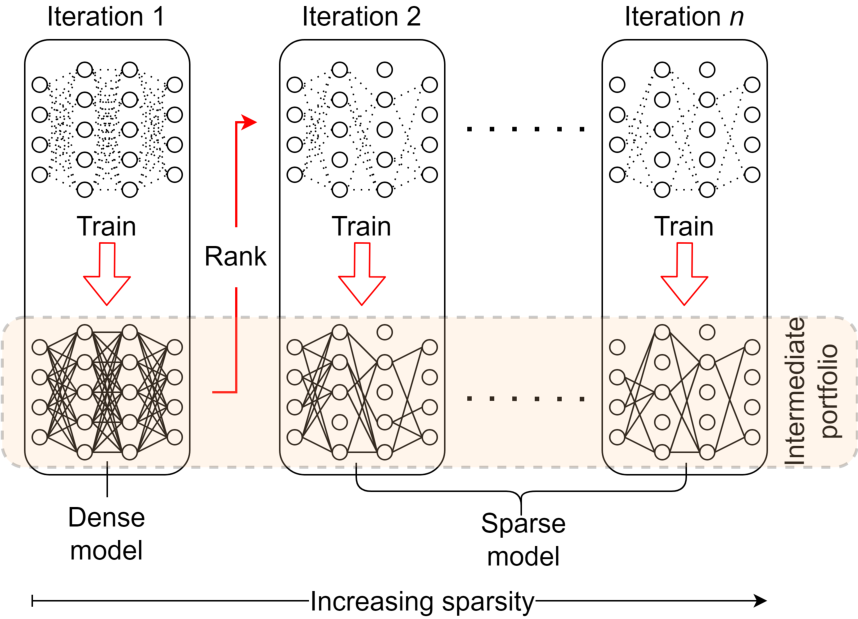}}
\caption{The unstructured pruning method incorporated in \DNNShifter uses the combined approach of repetitive training and model ranking between training iterations.}
\label{training}
\end{figure}

\DNNShifter implements a modified version of the Open Lottery Ticket Hypothesis (OpenLTH) framework\footnote{\url{https://github.com/facebookresearch/open_lth}}. No modifications were made to the training process. Instead, \DNNShifter adds the structured pruning and model switching phases (Phase 2 and Phase 3) which will be discussed later. The Lottery Ticket Hypothesis (LTH) articulates that highly accurate sparse models can be obtained from dense models~\cite{b16} (shown in Figure~\ref{training}) and is underpinned by Iterative Magnitude Pruning (IMP) with weight rewinding~\cite{frankle2019stabilizing} method that \DNNShifter also employs. IMP with rewinding is chosen since it performs well across all available models and datasets. Alternatives, such as SynFlow~\cite{b20}, only perform well for specific models or datasets~\cite{b17}. 

Figure~\ref{training} illustrates Phase 1 of \DNNShifter to generate sparse models using LTH. The model is trained and then ranked by the IMP algorithm in each iteration. The resulting sparse model from each iteration is saved into an intermediate portfolio of sparse models that will be pruned in the next phase. The sparse model from each iteration is provided as input for the next iteration. Model sparsity increases with the number of iterations up to a user-defined limit.

\subsection{Phase 2: Converting sparse models to pruned models}\label{phase2}

In this phase, the $n$ sparse models from the intermediate portfolio are converted into $m$ pruned models using structured pruning. This phase consists of \bailey{six processing modules} that pre-process sparse models, identify prunable data structures within each sparse model, generates plans for pruning, and then use structured pruning to generate pruned models. Each sparse model is processed from the intermediate portfolio to produce a final portfolio. \bailey{Each module of this phase is detailed below:}

\subsubsection*{\bailey{Module 1 - Model Pre-Processor}} 
This pre-processing module simplifies the DNN model architecture by fusing the convolution and batch normalisation layers~\cite{b10}. This fusion removes batch normalisation parameters, thereby reducing the complexity of generating a pruning plan for the model, as only convolutional layers and their dependants must be considered (further discussed in Module 3).

\subsubsection*{\bailey{Module 2 - Sparsity Analyser}} 
This module builds on the method illustrated in Figure~\ref{ussp} that identifies convolutional kernels with entirely zero values. However, these kernels cannot be removed without further planning since a DNN's architecture does not naturally lend itself to the removal of kernels alone. Instead, a kernel can be removed if all kernels in a channel can be removed. To this end, channels that have all kernels with zero values are further indexed to create prunable convolutional channels.

\subsubsection*{\bailey{Module 3 - Prune Planner}} 
In the existing literature, convolution channels are removed from a model iteratively to minimise accuracy loss. However, this is inefficient for two reasons. Firstly, pruning channels is computationally intensive since entire convolutional layers comprising large multi-dimensional parameter arrays will be rebuilt when prunable channels are removed. Secondly, each prunable channel depends on the channel of the next convolutional layer, which will also be rebuilt. Therefore, pruning sequentially incurs overheads. \DNNShifter breaks this dependency and removes all prunable channels at the same time. This is \bailey{achieved by the prune planner module,} which creates a concurrent data structure of prunable channel indices. 

This module uses Algorithm~\ref{alg:prune_planner} where each zero channel $c_{zero}$ (channels with all weights set to zero) in a set of all zero channels $C_{Zero}$ (indexed in Module 2) is mapped to a convolutional layer $L_n$ where $0 \leq n < D_{conv}$ (model convolutional layer depth). Each convolutional layer receives two sets of zero channels. The first set, $C_{in}$ is the set of prunable \textit{out} channels from the previous convolutional layer $L_{n-1}$: these indices correspond to the prunable \textit{in} channels of $L_n$. The second set, $C_{out}$, is the prunable \textit{out} channels of $L_n$. When $n = 0$ (first convolutional layer), there is no $C_{in}$. Therefore this layer receives an empty set. The returned prune plan ($(C_{in}, C_{out})$) contains all zero channels that are to be pruned in Module 4 for a given convolutional layer.
\begin{algorithm}[tp]
\algsetup{linenosize=\tiny}
\SetAlgoLined
\KwData{Set of all zero channels $C_{Zero}$,\\  Convolutional Layer $L_n$}
\KwResult{Prunable channels $(C_{in}, C_{out})$ in $L_n$}
$C_{in} \gets \{\}$ \\
$C_{out} \gets \{ \; c_{zero}\; | \; c_{zero} \in (\;L_n\; \cap \; C_{Zero} \;)\}$ \\
\uIf{($n > 0$)}{
    $C_{in} \gets \{ \; c_{zero} \; | \; c_{zero} \in (\;L_{n-1}\; \cap \; C_{Zero}\;)\}$
}
\Return{} $(C_{in}, C_{out})$
\caption{\DNNShifter Prune Planning}
\label{alg:prune_planner}
\end{algorithm}

\subsubsection*{\bailey{Module 4 - Model Pruner}} 
The pruning plan from Module 3 is used to prune a sparse model from the intermediate portfolio in real time. This module executes the pruning plan by rebuilding each convolutional layer without the prunable channels and the biases of the channels. As all prunable channels are made available from Module 3, this module prunes all in/out channels in a single batch operation, significantly reducing computational overhead and enabling real-time pruning. After prune planning, this module is executed in parallel to concurrently prune each convolutional layer, forming a series of pruned layers $L^{\prime}$ that replaces the original unpruned layers $L$.

This module uses in Algorithm~\ref{alg:batch_pruner} where a pruned layer $L_{n}^{\prime}$ is created with the smaller channel size $|C_{in}^{\prime}|$ and $|C_{out}^{\prime}|$ (Lines 1-3). Afterwards, the pruned set of remaining channels and bias are transferred from $L_n$ to $L_{n}^{\prime}$ (Lines 4-7).

\begin{algorithm}[tp]
\algsetup{linenosize=\tiny}
\SetAlgoLined
\KwData{Prunable channel indices $(C_{in}, C_{out})$ in $L_n$}
\KwResult{Pruned convolutional layer $L_{n}^{\prime}$}
$|C_{in}^{\prime}| \gets |L_{n}(C_{in})| - |C_{in}|$ \\
$|C_{out}^{\prime}| \gets |L_{n}(C_{out})| - |C_{out}|$ \\
$L_{n}^{\prime} \gets $ create new layer($|C_{in}^{\prime}|$, $|C_{out}^{\prime}|$) \\
$L_{n}^{\prime}(C_{out}) \gets $ $L_{n}(C_{out}) \setminus C_{out}$\\ 
$L_{n}^{\prime}(bias) \gets $ $L_{n}(bias) \setminus C_{out}$\\ 
\uIf{($n > 0$)}{
    $L_{n}^{\prime}(C_{in}) \gets $ $L_{n}(C_{in}) \setminus C_{in}$\\
}
\Return{} $L_{n}^{\prime}$
\caption{\DNNShifter Batch Pruning}
\label{alg:batch_pruner}
\end{algorithm}

\subsubsection*{\bailey{Module 5 - Model Profiler}}
This module benchmarks the pruned model to obtain metrics: accuracy, inference latency, model size, and the maximum memory required to run the model. This is achieved using a test dataset. The metrics relevant to each model are stored as metadata for the next module for selecting a suitable pruned model from a portfolio.

\subsubsection*{\bailey{Module 6 - Portfolio Post-Processor}} 
A portfolio of $n$ pruned models is generated. This module refines the portfolio to eventually only include $m$ pruned models ($m \leq n$) with distinct performance characteristics (pruned models with similar characteristics are removed).
\subsection{Phase 3: Further compression and model switching}
\label{phase3}
A portfolio of production-quality DNN models is trained in the first phase and then compressed via structured pruning in the second phase. In this third phase, models are further compressed while not being used (inactive). \DNNShifter encodes the portfolio of pruned models into a significantly smaller package before deploying it to the storage of the target device using the lossless DEFLATE~\cite{rfc1951} compression algorithm. On application initialisation, \DNNShifter loads the entire portfolio into memory, and then one model is activated (inflated) to enable inference. Encoding models in this manner (deflating) is possible since zero weights repeat in highly sparse DNN models obtained from training. However, out-of-the-box production quality models are dense (most of their weights are not set to zero). Therefore, applying this compression algorithm to dense models will not provide any benefit. Each module of this phase is detailed below:

\subsubsection*{\bailey{Module 7 - Model Deflater}} This module sorts the model portfolio by model size (a proxy for performance characteristics), then shrinks the entire model portfolio into a smaller, sorted, and easily deployable package using DEFLATE before it is transferred to the target devices.

\subsubsection*{\bailey{Module 8 - Application Initialser}} This module loads the entire portfolio of deflated models into device memory. First, it selects a model with the median model size. \bailey{Then, this model is decompressed in memory to enable application inference (we refer to this as inflation). Note that the inflated model is a pruned model variant from Phase 2. It is smaller and faster for inference than an equivalent dense model (Figure 6).}

\subsubsection*{\bailey{Module 9 - Model Switcher}} \bailey{The available memory and CPU load may vary due to the number of running applications and the workload of each application on the device. For example, inference performance metrics, such as queries per second (QPS), may vary over time for an application~\cite{infaas}. During a low load on the edge device, a higher QPS can be achieved, during which time a larger model from the portfolio can be decompressed in the device memory (inflation); the larger model will improve inference accuracy. Inversely, a decreasing QPS suggests a high load and a smaller model from the portfolio is inflated to improve inference latency performance. \DNNShifter does not require searching the entire portfolio to switch between models. Instead, this module selects the next or previous model from the portfolio depending on the QPS trend. Therefore, model switching can be rapidly obtained with minimum overheads.}

\section{Experiments}
\label{sec:experimentalstudies}
This section first presents the experimental testbed and baseline models in Section~\ref{subsec:exp_setup1} and then considers three key aspects of \DNNShifterr: 

(1) The time to generate a portfolio of models for addressing Challenge 1 (Phase 1 of \DNNShifterr). We will compare against state-of-the-art NAS methods, namely DARTS~\cite{liu2018darts}, RepVGG~\cite{repvgg}\bailey{, and PreNAS~\cite{prenas}}, for evaluating this. We will demonstrate in Section~\ref{subsec:new_exp1} that \DNNShifter will generate a portfolio faster and more efficiently.

(2) The accuracy achieved and the time taken for inference by the models for addressing Challenge 2 (Phase 2 of \DNNShifterr). 
Two categories of pruning algorithms, unstructured and structured, are considered here. The unstructured pruning algorithms considered are random pruning, magnitude pruning~\cite{obd}, SynFlow~\cite{b20}\bailey{, and NTK-SAP~\cite{ntksap}.} The structured pruning algorithms considered include similarities-aware~\cite{classical_struc}, $l^{1}$ norm~\cite{classical_struc}, EasiEdge~\cite{easiedge}, \bailey{and ProsPr~\cite{prospr}}. 
We will demonstrate in Section~\ref{subsec:new_exp2a} and Section~\ref{subsec:new_exp2b} that \DNNShifter obtains better accuracy and an improved inference speedup for the pruned models compared to unstructured pruning methods. It will also be demonstrated in Section~\ref{subsec:new_exp3} that when compared to structured pruning methods, the pruned models obtained from \DNNShifter have better accuracy for a desired model size.  
We will show in Section~\ref{subsec:new_exp4} that \DNNShifter has overheads that are multiple magnitudes lower than structured pruning methods.

(3) The overheads for dynamically switching a model in memory for addressing Challenge 3 (Phase 3 of \DNNShifterr). We will demonstrate in Section~\ref{subsec:results_phase3} that compared to model switching approaches, such as Model Ensemble~\cite{ensemble} and Dynamic once-for-all (Dynamic-OFA)~\cite{dynamic-ofa}, \DNNShifter has lower model switching overheads and memory utilisation. 

\subsection{Experimental setup} 
\label{subsec:exp_setup1}
Two production DNN models trained on the CIFAR-10~\cite{cifar10} and Tiny ImageNet~\cite{le2015tiny} dataset are considered. First is VGG-16~\cite{b3} trained on CIFAR-10, which represents a feedforward DNN model. Second is ResNet-50~\cite{resnet} trained on Tiny ImageNet, which is a more complex branching DNN model. 
\begin{table}[hbt]
\caption{\bailey{Baseline results and training hyperparameters for a production quality dense VGG-16 and ResNet-50 models.}}
\label{tab:exp1vanillamodels}
\centering
\begin{tabular}{lrr}
\hline
                & \multicolumn{1}{r}{VGG-16} & \multicolumn{1}{r}{ResNet-50}     \\ \hline
Dataset         & CIFAR-10                   & Tiny ImageNet \\
Params. (M)     & 14.72                      & 25.56                             \\
Size (MB)       & 56.2                       & 97.8                              \\
Top-1 Acc. (\%) & 93.21 \bailey{± 0.06}                      & 54.99 \bailey{± 0.14}                                \\
\bailey{Epochs}          & \bailey{160}                        & \bailey{200}                               \\
\bailey{Batch Size}      & \bailey{128}                        & \bailey{256}                               \\
\bailey{Learning Rate}   & \bailey{0.1}                        & \bailey{0.2}                               \\
\bailey{Milestone Steps}\tablefootnote{\bailey{Learning rate drops by a factor of gamma, 0.1, at each milestone step. Both models use the SGD optimiser, momentum of 0.9 and weight decay of 0.0001.}} & \bailey{80, 120}                    & \bailey{100, 150}                          \\ \hline
\end{tabular}
\end{table}

Table~\ref{tab:exp1vanillamodels} presents the baseline results \bailey{and hyperparameters}:

\textbf{Models, Datasets, and Hyperparameters -} VGG-16 is the OpenLTH configuration that has \bailey{one linear layer}~\cite{b16}, and ResNet-50 is the default ImageNet configuration~\cite{resnet}. CIFAR-10 consists of 50,000 training images and 10,000 test images divided equally across 10 classes. Tiny ImageNet is a subset of ImageNet consisting of 100,000 training images and 10,000 test images divided equally across 200 classes. \bailey{Tiny ImageNet results are reported for both Top-1 and Top-5 as recommended by model pruning literature~\cite{bg2}.} The baseline results were obtained using the training routine from OpenLTH\footnote{Using Python 3.8.10, torch 1.13.0+cu116, and torchvision 0.14.0+cu116.} as defined in Section~\ref{overview}.

\textbf{\bailey{Testbed} -} We use an AMD EPYC 7713P 64-core CPU and Nvidia RTX A6000 GPU to train the models, as such resources are representative of those in a cloud data centre. \bailey{Model inference and runtime switching is} carried out with an Intel i7-9750H \bailey{6-core} CPU and an Nvidia RTX 2080 (Max-Q) GPU comparable to an edge server that may be used in a production setting.

\bailey{\textbf{Trial Counts and Reporting Methods -} All DNN training methodologies and experiments were conducted a minimum of three times, except for those in Section~\ref{subsec:new_exp1}. In Section~\ref{subsec:new_exp1}, each NAS approach was executed only once due to computational and time constraints. Unless otherwise specified, model performance indicators like accuracy, memory usage, and latency are presented in tables and figures as the mean from all trials accompanied by confidence intervals spanning one standard deviation. In addition, where possible, experiments are carried out across 8 different compression ratios (2, 4, 8, 16, 32, 64, 128, 256).}

\subsection{Model training and portfolio generation (Phase 1)}

\label{subsec:new_exp1}

\begin{table*}[t]
	\caption{\bailey{Comparing the search efficiency} of \DNNShifter against NAS methods when generating a portfolio of model variants. \bailey{Mean Top-1 accuracy and standard deviation of the optimal model variant portfolio is recorded for the CIFAR-10 dataset.} }
	\label{tab:new_exp1trainingtime}
	\centering
\begin{tabular}{lrrrrrrcrr}
\hline
           & \multicolumn{1}{l}{}                  & \multicolumn{3}{c}{Model Params. (M)}                                                & \multicolumn{1}{l}{}                    & \multicolumn{1}{l}{}         & \multicolumn{1}{l}{}              & \multicolumn{1}{l}{}                          & \multicolumn{1}{l}{}                  \\ \cline{3-5}
           & \multicolumn{1}{c}{\makecell{\# Model \\ Variants}} & \multicolumn{1}{c}{Min.} & \multicolumn{1}{c}{Avg.} & \multicolumn{1}{l}{Max.} & \multicolumn{1}{l}{\makecell{Params. \\Trained (M)}} & \multicolumn{1}{c}{GPU-days} & \multicolumn{1}{c}{Acc. (\%)} & \multicolumn{1}{c}{\makecell{\# Optimal\\ Variants}} & \multicolumn{1}{l}{\makecell{Search \\Efficiency (\%)}} \\ \hline

\bailey{PreNAS~\cite{prenas}} & \bailey{1770} & \bailey{41.44} & \bailey{77.64} & \bailey{102.78} & \bailey{543.5}& \bailey{9.11} & \bailey{96.96 ± 1.48} & \bailey{7} & \bailey{0.39}\\
RepVGG~\cite{repvgg}     & 228                                     & 8.30                        & 48.81                        & 110.96                        & 12329.0                                       & 26.09                            & 94.96 ± 0.16                                 & 13                                             & 5.70                                     \\
DARTS~\cite{liu2018darts}      & 50                                     & 3.30                        & 3.30                        & 3.30                        & 165.0                                      & 1.77                            & 74.01 ± 16.9                                 & 3                                             & 8.00                                     \\ \hline
\DNNShifter & 9                                     & 0.05                        & 3.32                        & 14.72                        & 132.2                                       & 0.28                            & 93.25 ± 0.66                                 &               4                               & 44.44                                     \\\hline
\end{tabular}
\end{table*}

This study will demonstrate that \DNNShifter will generate a portfolio of models from a base architecture with a higher search efficiency than comparable NAS methods. Search efficiency is the percentage of optimal model variants in the portfolio over the total number of searched variants. An optimal model variant is one that is not outperformed on all performance metrics by another variant and is obtained using Pareto optimality. The performance metrics considered in this article are model size, inference latency, and model accuracy. For example, training a single model that reaches an adequate accuracy has a search efficiency of 100\%. However, if training \bailey{occurs} $N$ times, then the model with the highest accuracy from the $N$ training rounds is optimal, but search efficiency drops to $100/N$\%.

\DNNShifter creates a model portfolio by iteratively pruning the largest model variant into progressively smaller variants (discussed in Section~\ref{phase1}). Each pruning iteration is equivalent to searching the model architecture once for a variant. We compare \DNNShifter against \bailey{three} different NAS\bailey{-based} methods, which search a model architecture for optimal model variants. 

The first is DARTS, an accelerated NAS method that generates model variants from a continuous search space via gradient descent. Compared to older NAS methods such as NasNet, DARTS is 500x faster. In addition, DARTS is a NAS approach that automatically generates a portfolio of models.

The second is RepVGG, which employs a family of VGG-like model variants from a set of discrete hyperparameters that scale various model architecture properties. 
In total, 228 model variants are individually trained to identify the optimal set of model variants based on hyperparameters presented in RepVGG literature~\cite{repvgg}.

\bailey{The third is PreNAS, a modern NAS that generates models using the emerging vision transformer model architecture~\cite{vt}. PreNAS is a one-shot NAS that decides on a set of optimal model variants and only trains those candidates, significantly reducing computational requirements.}

\DNNShifter has one hyperparameter, $n$, which specifies how many model variants should be generated where each variant is twice as compressed as the previous. For example, $n=8$ generates a portfolio of model variants up to the compression ratio 256 ($2^{n}$). The first variant is the original dense model with no sparsity.

Table~\ref{tab:new_exp1trainingtime} contrasts \DNNShifter against DARTS, RepVGG, \bailey{and PreNAS}. \DNNShifter generates 4 optimal model variants out of a portfolio of 9, resulting in a search efficiency of 44.44\%. This is more efficient \bailey{than the NAS-based methods}. The number of parameters trained using DARTS is divided across a more extensive portfolio of models. These models are not sufficiently diverse, resulting in a low search efficiency of DARTS. The DARTS search method requires over 6 hours to create the portfolio. Then, each variant is individually trained and evaluated, totalling a training time \bailey{that is 6x longer} than \DNNShifterr. RepVGG \bailey{and PreNAS} achieve a higher model accuracy than \DNNShifterr, but each model variant is \bailey{up to one order of magnitude larger in parameter count.} As this study evaluates training time as a function of parameter count, the trend that is seen in Table~\ref{tab:new_exp1trainingtime} generalises for all model architectures of different sizes. 

\begin{tcolorbox}[
    width=0.485\textwidth,
    enhanced,
    colframe=black!50!black,
    colback=gray!5,]
\bailey{
\textbf{Observation 1:}
The \DNNShifter method for generating a portfolio of models via iterative pruning is more resource and time-efficient than NAS-based methods.}
\end{tcolorbox}
\subsection{Performance of sparse and pruned models (Phase 2)}
\label{subsec:results_phase2}
This study will demonstrate that \DNNShifter produces pruned models of the same or better accuracy than other unstructured (Section~\ref{subsec:new_exp2a}) and structured (Section~\ref{subsec:new_exp3}) pruning methods. In addition, it is demonstrated that the pruned models generated by \DNNShifter are smaller and faster than sparse models, which is quantified for various compression ratios (Section~~\ref{subsec:new_exp2b}).

\subsubsection{Comparing accuracy against sparse models}
\label{subsec:new_exp2a}
We will first contrast the choice of the unstructured pruning method in \DNNShifter against other unstructured pruning methods. Unstructured pruning methods produce a sparse model variant where parameters are set to zero. In this paper, a sparse model with a compression ratio of $C$ has for every $C$ parameters, $C-1$ parameters set to zero; this is also presented in the literature~\cite{b20, bg2}. As seen in Section~\ref{phase1}, \DNNShifter utilises IMP with rewinding as its unstructured pruning method. This study evaluates \DNNShifter against random pruning, which is a naive method, magnitude pruning, SynFlow\bailey{, and NTK-SAP}. For each method, the baseline models in Table~\ref{tab:exp1vanillamodels} are iteratively pruned up to eight times where a compression ratio of $2^{n}$ is achieved per iteration $n$ as described in Section \ref{subsec:new_exp1}.

\begin{figure*}[!t]
    \centering
\begin{tikzpicture}
    \begin{groupplot}[group style={group size= 3 by 1, vertical sep=2cm, horizontal sep=1.5cm},height=5cm,width=6cm, axis background/.style={fill=gray!10}]
        \nextgroupplot[title=VGG-16 (CIFAR-10), ylabel = Top-1 Accuracy (\%), xmin=1, xmax = 256, ymin=0, xtick={32,64,128,256}]
        
        \addplot[line width=0.25mm, blue, mark=*, mark size=1.5pt, error bars/.cd, y dir=both, y explicit,] coordinates {
        (1, 93.21)
        (2, 93.42)
        (4, 93.51)
        (8, 93.66)
        (16, 93.71)
        (32, 93.61)
        (64, 93.64)
        (128, 92.82) +=(128,0) -=(128,0)
        (256, 91.64)
        };\label{plots:plot9}
        
        \addplot[smooth, line width=0.25mm,red, mark=square*, mark size=1.5pt,] coordinates {
        (1, 93.21)
        (2, 93.49)
        (4, 92.74)
        (8, 92.67)
        (16, 92.02)
        (32, 91.60)
        (64, 90.70)
        (128, 75.25)
        (256, 10.00)
        };\label{plots:plot10}
        \addplot[smooth, line width=0.25mm,green, mark=triangle*, mark size=1.5pt,] coordinates {
        (1, 93.21)
        (2, 93.45)
        (4, 93.26)
        (8, 93.05)
        (16, 92.47)
        (32, 91.59)
        (64, 90.85)
        (128, 89.00)
        (256, 87.00)
        };\label{plots:plot11}
        \addplot[smooth, line width=0.25mm,orange, mark=diamond*, mark size=1.5pt,] coordinates {
        (1, 93.21)
        (2, 92.73)
        (4, 92.39)
        (8, 91.06)
        (16, 89.56)
        (32, 87.03)
        (64, 83.81)
        (128, 79.24)
        (256, 62.12)
        };\label{plots:plot12}
        \addplot[smooth, line width=0.25mm,purple!80!white, mark=asterisk, mark size=1.5pt, error bars/.cd, y dir=both, y explicit,] coordinates {
        (1, 93.25) +=(1,0.06) -=(1,0.06)
        (2, 93.47) +=(2,0.01) -=(2,0.01)
        (4, 93.40) +=(4,0.08) -=(4,0.08)
        (8, 93.12) +=(8,0.21) -=(8,0.21)
        (16, 92.52) +=(16,0.28) -=(16,0.28)
        (32, 92.22) +=(32,0.23) -=(32,0.23)
        (64, 91.25) +=(64,0.05) -=(64,0.05)
        (128, 89.78) +=(128,0.13) -=(128,0.13)
        (256, 88.06) +=(256,0.08) -=(256,0.08)
        }
        ;\label{plots:plot13}
        \draw[dashed] (1,93.21) -- (256, 93.21);
        
        \nextgroupplot[title=ResNet-50 (Tiny ImageNet), xlabel= Compression Ratio, ylabel = Top-1 Accuracy (\%),
      xmin=1, xmax = 256, ymin=0, xtick={32,64,128,256}]
        \addplot[smooth, line width=0.25mm, blue, mark=*, mark size=1.5pt, error bars/.cd, y dir=both, y explicit,] coordinates {
        (1, 54.99) +=(1,0.0) -=(1,0.0)
        (2, 55.31) +=(2,0.137) -=(2,0.137)
        (4, 55.78) +=(4,0.27) -=(4,0.27)
        (8, 56.04) +=(8,0.156) -=(8,0.156)
        (16, 55.28) +=(16,0.267) -=(16,0.267)
        (32, 53.62) +=(32,0.321) -=(32,0.321)
        (64, 50.73) +=(64,0.411) -=(64,0.411)
        (128, 46.40) +=(128,0.197) -=(128,0.197)
        (256, 40.65) +=(256,0.0) -=(256,0.0)
        };\label{plots:plot9}
        \addplot[smooth, line width=0.25mm,red, mark=square*, mark size=1.5pt, error bars/.cd, y dir=both, y explicit,] coordinates {
        (1, 54.99) +=(1,0.0) -=(1,0.0)
        (2, 55.02) +=(2,0.137) -=(2,0.137)
        (4, 53.98) +=(4,0.27) -=(4,0.27)
        (8, 51.94) +=(8,0.156) -=(8,0.156)
        (16, 50.17) +=(16,0.267) -=(16,0.267)
        (32, 19.13) +=(32,0.321) -=(32,0.321)
        (64, 5.5) +=(64,0.411) -=(64,0.411)
        (128, 0.5) +=(128,0.197) -=(128,0.197)
        (256, 0.5) +=(256,0.0) -=(256,0.0)
        };\label{plots:plot10}
        \addplot[smooth, line width=0.25mm,green, mark=triangle*, mark size=1.5pt, error bars/.cd, y dir=both, y explicit,] coordinates {
        (1, 54.99) +=(1,0.0) -=(1,0.0)
        (2, 54.78) +=(2,0.137) -=(2,0.137)
        (4, 54.60) +=(4,0.27) -=(4,0.27)
        (8, 54.3) +=(8,0.156) -=(8,0.156)
        (16, 52.15) +=(16,0.267) -=(16,0.267)
        (32, 49.6) +=(32,0.321) -=(32,0.321)
        (64, 42.8) +=(64,0.411) -=(64,0.411)
        (128, 35.82) +=(128,0.197) -=(128,0.197)
        (256, 24.57) +=(256,0.0) -=(256,0.0)
        };\label{plots:plot11}
        \addplot[smooth, line width=0.25mm,orange, mark=diamond*, mark size=1.5pt, error bars/.cd, y dir=both, y explicit,] coordinates {
        (1, 54.99) +=(1,0.0) -=(1,0.0)
        (2, 54.12) +=(2,0.137) -=(2,0.137)
        (4, 50.93) +=(4,0.27) -=(4,0.27)
        (8, 48.93) +=(8,0.156) -=(8,0.156)
        (16, 44.9) +=(16,0.267) -=(16,0.267)
        (32, 41.33) +=(32,0.321) -=(32,0.321)
        (64, 35.46) +=(64,0.411) -=(64,0.411)
        (128, 24.78) +=(128,0.197) -=(128,0.197)
        (256, 13.63) +=(256,0.0) -=(256,0.0)
        };\label{plots:plot12}
        \addplot[smooth, line width=0.25mm,purple!80!white, mark=asterisk, mark size=1.5pt, error bars/.cd, y dir=both, y explicit,] coordinates {
        (1, 54.99) +=(1,0.0) -=(1,0.0)
        (2, 55.37) +=(2,0.137) -=(2,0.137)
        (4, 55.25) +=(4,0.27) -=(4,0.27)
        (8, 54.62) +=(8,0.156) -=(8,0.156)
        (16, 54.2) +=(16,0.267) -=(16,0.267)
        (32, 52.97) +=(32,0.321) -=(32,0.321)
        (64, 46.62) +=(64,0.411) -=(64,0.411)
        (128, 1.183) +=(128,0.197) -=(128,0.197)
        (256, 0.5) +=(256,0.0) -=(256,0.0)
        };\label{plots:plot13}
        \draw[dashed] (1,54.99) -- (256, 54.99);

        \nextgroupplot[title=ResNet-50 (Tiny ImageNet), ylabel = Top-5 Accuracy (\%), xmin=1, xmax = 256, ymin=0, xtick={32,64,128,256}]

        \addplot[smooth, line width=0.25mm, blue, mark=*, mark size=1.5pt, error bars/.cd, y dir=both, y explicit,] coordinates {
        (1, 77.6)
        (2, 78.02)
        (4, 78)
        (8, 78.87)
        (16, 78.52)
        (32, 78.17) +=(32,0.257) -=(32,0.257)
        (64, 75.89) +=(64,0.3122) -=(64,0.3122)
        (128, 72.47) +=(128,0.061) -=(128,0.061)
        (256, 67.64)
        };\label{plots:plot9}
        \addplot[smooth, line width=0.25mm,red, mark=square*, mark size=1.5pt, error bars/.cd, y dir=both, y explicit,] coordinates {
        (1, 77.6)
        (2, 77.62)
        (4, 77.2)
        (8, 75.88)
        (16, 75.17)
        (32, 29.59) +=(32,0.257) -=(32,0.257)
        (64, 12.5) +=(64,0.3122) -=(64,0.3122)
        (128, 2.5) +=(128,0.061) -=(128,0.061)
        (256, 2.5)
        };\label{plots:plot10}
        \addplot[smooth, line width=0.25mm,green, mark=triangle*, mark size=1.5pt, error bars/.cd, y dir=both, y explicit,] coordinates {
        (1, 77.6)
        (2, 77.56)
        (4, 77.8)
        (8, 78.82)
        (16, 77.56)
        (32, 75.68) +=(32,0.257) -=(32,0.257)
        (64, 69.67) +=(64,0.3122) -=(64,0.3122)
        (128, 62.69) +=(128,0.061) -=(128,0.061)
        (256, 45.83)
        };\label{plots:plot11}
        \addplot[smooth, line width=0.25mm, orange, mark=diamond*, mark size=1.5pt, error bars/.cd, y dir=both, y explicit,] coordinates {
        (1, 77.6)
        (2, 77.08)
        (4, 75.4)
        (8, 73.44)
        (16, 70.55)
        (32, 67.49) +=(32,0.257) -=(32,0.257)
        (64, 62.11) +=(64,0.3122) -=(64,0.3122)
        (128, 49.37) +=(128,0.061) -=(128,0.061)
        (256, 33.79)
        };\label{plots:plot12}
        \addplot[smooth, line width=0.25mm, purple!80!white, mark=asterisk, mark size=1.5pt, error bars/.cd, y dir=both, y explicit,] coordinates {
        (1, 77.6) +=(1,0.0) -=(1,0.0)
        (2, 78.09) +=(2,0.065) -=(2,0.065)
        (4, 78.17) +=(4,0.298) -=(4,0.298)
        (8, 78.51) +=(8,0.17) -=(8,0.17)
        (16, 78.44) +=(16,0.035) -=(16,0.035)
        (32, 77.51) +=(32,0.257) -=(32,0.257)
        (64, 70.95) +=(64,0.3122) -=(64,0.3122)
        (128, 10) +=(128,0.061) -=(128,0.061)
        (256, 2.5) +=(256,0.0) -=(256,0.0)
        };\label{plots:plot13}
        \draw[dashed] (1,77.6) -- (256, 77.6);
\coordinate (bot) at (rel axis cs:1,0);
    \end{groupplot}
\path (current bounding box.north)--
      coordinate(legendpos)
      (current bounding box.north);
\matrix[
    matrix of nodes,
    anchor=south,
    draw,
    inner sep=0.1em,
    draw
  ]at([yshift=1ex]legendpos)
  {
    \ref{plots:plot9}& \DNNShifterr &[5pt]
    \ref{plots:plot10}& Magnitude&[5pt]
    \ref{plots:plot11}& SynFlow&[5pt]
    \ref{plots:plot13}& NTK-SAP&[5pt]
    \ref{plots:plot12}& Random\\};
\end{tikzpicture}
\caption{\bailey{Accuracy of unstructured pruning in \DNNShifter against other methods as compression ratio increases; dashed line is baseline model accuracy.}}
\label{new_exp2a}
\end{figure*}
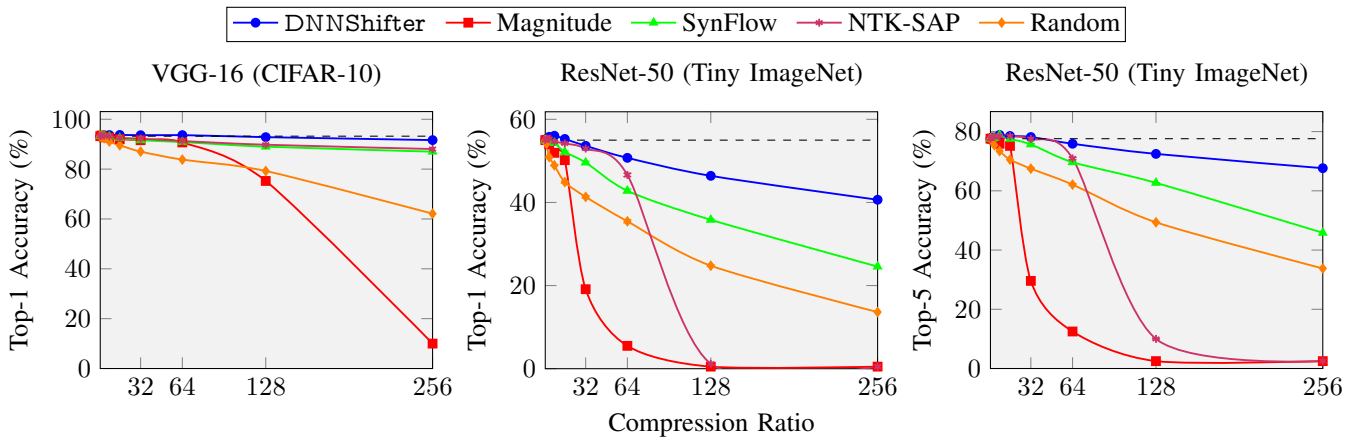

Figure~\ref{new_exp2a} shows the change in test accuracy as the compression ratio increases for VGG-16 and Resnet-50. For all methods, accuracy decreases as the compression ratio increases. However, the rate of decline varies per method, whereas \DNNShifter maintains the highest accuracy in all cases. For example, for VGG-16 on CIFAR-10 at a compression ratio of 256, \DNNShifter accuracy dropped by 1.57\%, whereas SynFlow\bailey{, NTK-SAP,} and Random pruning dropped by 6.21\%\bailey{, 5.15\%,} and 31.09\%, respectively. Magnitude pruning compromises almost all of its accuracy by a compression ratio of 256. Magnitude pruning causes the model to become unusable at high compression ratios due to layer collapse (when an entire layer is pruned)~\cite{b20}. However, SynFlow was designed to avoid layer collapse and maintain usable accuracy at high compression ratios. \DNNShifter also does not encounter layer collapse as it uses the rewinding approach from IMP that stabilises the sparse model~\cite{frankle2019stabilizing}.

For ResNet-50 on Tiny ImageNet, for top-1 and top-5 test accuracy, \DNNShifter accuracy reduction is less than half of SynFlow at higher compression ratios. In contrast, magnitude pruning\bailey{, and NTK-SAP} undergo layer collapse after a compression ratio of 32\bailey{, and 64, respectively}.

\begin{tcolorbox}[
    width=0.485\textwidth,
    enhanced,
    colframe=black!50!black,
    colback=gray!5,]
\bailey{
\textbf{Observation 2:}
\DNNShifter preserves the highest accuracy for sparse models compared to existing methods. Furthermore, this enables \DNNShifter to generate sparse models at extreme compression ratios, providing the opportunity for structured pruning.}
\end{tcolorbox}

The benefits of using structured pruning \bailey{at extreme model sparsities} are explored \bailey{in the} next \bailey{subsection}.

\disablehyperlinks
\enablehyperlinks
\subsubsection{Comparing runtime performance against sparse models}
\label{subsec:new_exp2b}
Unstructured pruning methods do not provide inference acceleration or spatially reduce the model size at runtime. This is because the parameters are not spatially pruned but rather are merely made zero values in the model. 
\DNNShifter removes pruned parameters via structured pruning. This study highlights the performance benefits of using pruned models from \DNNShifter compared to the sparse models generated by unstructured pruning methods.

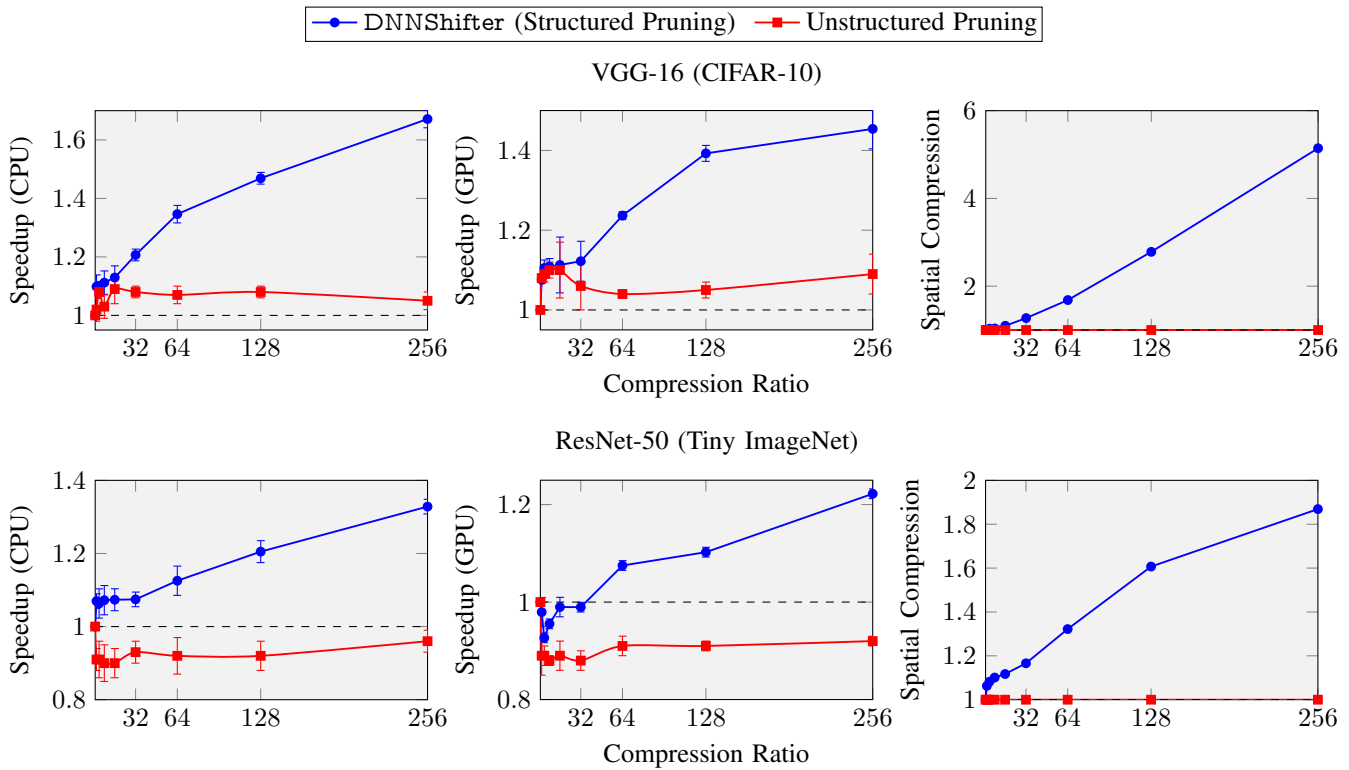
\begin{figure*}[!t]
    \centering
\begin{tikzpicture}
    \begin{groupplot}[group style={group size= 3 by 2, vertical sep=2cm, horizontal sep=1.5cm},height=4.5cm,width=6cm, axis background/.style={fill=gray!10}]
        \nextgroupplot[ylabel = Speedup (CPU),
      xmin=1, xmax = 256, ymin=0.95, ymax = 1.7, xtick={32,64,128,256}]
        \addplot[line width=0.25mm, blue, mark=*, mark size=1.5pt, error bars/.cd, y dir=both, y explicit,] coordinates {
        (1, 1) += (1,0.00) -=(1, 0.00)
(2,1.09839999999999) += (2,0.04) -=(2, 0.04)
(4,1.10139999999999) += (4,0.01) -=(4, 0.01)
(8,1.11199999999999) += (8,0.04) -=(8, 0.04)
(16,1.1294) += (16,0.04) -=(16, 0.04)
(32,1.2066) += (32,0.02) -=(32, 0.02)
(64,1.3462) += (64,0.03) -=(64, 0.03)
(128,1.46879999999999) += (128,0.02) -=(128, 0.02)
(256,1.67139999999999) += (256,0.03) -=(256, 0.03)
        };\label{plots:plot5}
        \addplot[smooth, line width=0.25mm,red, mark=square*, mark size=1.5pt, error bars/.cd, y dir=both, y explicit,] coordinates {
        (1, 1) += (1,0.00) -=(1, 0.00)
        (2,1.02) += (2,0.04) -=(2, 0.04)
        (4,1.08) += (4,0.01) -=(4, 0.01)
        (8,1.03) += (8,0.04) -=(8, 0.04)
        (16,1.09) += (16,0.05) -=(16, 0.05)
        (32,1.08) += (32,0.02) -=(32, 0.02)
        (64,1.07) += (64,0.03) -=(64, 0.03)
        (128,1.08) += (128,0.02) -=(128, 0.02)
        (256,1.05) += (256,0.03) -=(256, 0.03)
        };\label{plots:plot6}
        \draw[dashed] (1,1.0) -- (256, 1.0);

        \nextgroupplot[title=VGG-16 (CIFAR-10), xlabel= Compression Ratio, ylabel = Speedup (GPU), xmin=1, xmax = 256, ymin=0.95, ymax=1.5, xtick={32,64,128,256}]

        \addplot[line width=0.25mm, blue, mark=*, mark size=1.5pt, error bars/.cd, y dir=both, y explicit,] coordinates {
        (1, 1.) += (1,0.00) -=(1, 0.00)
(2,1.07579999999999) += (2,0.01) -=(2, 0.01)
(4,1.10519999999997) += (4,0.02) -=(4, 0.02)
(8,1.10879999999999) += (8,0.02) -=(8, 0.02)
(16,1.11279999999999) += (16,0.07) -=(16, 0.07)
(32,1.12219999999999) += (32,0.05) -=(32, 0.05)
(64,1.23659999999999) += (64,0.01) -=(64, 0.01)
(128,1.3926) += (128,0.02) -=(128, 0.02)
(256,1.45399999999999) += (256,0.05) -=(256, 0.05)
        };\label{plots:plot5}
        \addplot[smooth, line width=0.25mm,red, mark=square*, mark size=1.5pt, error bars/.cd, y dir=both, y explicit,] coordinates {
        (1, 1) += (1,0.00) -=(1, 0.00)
        (2,1.08) += (2,0.01) -=(2, 0.01)
        (4,1.09) += (4,0.02) -=(4, 0.02)
        (8,1.10) += (8,0.02) -=(8, 0.02)
        (16,1.10) += (16,0.07) -=(16, 0.07)
        (32,1.06) += (32,0.06) -=(32, 0.06)
        (64,1.04) += (64,0.01) -=(64, 0.01)
        (128,1.05) += (128,0.02) -=(128, 0.02)
        (256,1.09) += (256,0.05) -=(256, 0.05)
        };\label{plots:plot6}
        \draw[dashed] (1,1) -- (256, 1);

        \nextgroupplot[ylabel = Spatial Compression, xmin=1, xmax = 256, ymin=1.0, ymax=6.0, xtick={32,64,128,256}]
        \addplot[line width=0.25mm, blue, mark=*, mark size=1.5pt, error bars/.cd, y dir=both, y explicit,] coordinates {
        (1, 1)
        (2, 1.001)
        (4, 1.03)
        (8, 1.035)
        (16, 1.097)
        (32, 1.272)
        (64, 1.683)
        (128, 2.782)
        (256, 5.144)
        };\label{plots:plot5}
        \addplot[smooth, line width=0.25mm,red, mark=square*, mark size=1.5pt,] coordinates {
        (1, 1)
        (2, 1)
        (4, 1)
        (8, 1)
        (16, 1)
        (32, 1)
        (64, 1)
        (128, 1)
        (256, 1)
        };\label{plots:plot6}
        \draw[dashed] (1,1) -- (256, 1);

        \nextgroupplot[ ylabel = Speedup (CPU),
      xmin=1, xmax = 256, ymin=0.8, ymax=1.4, xtick={32,64,128,256}]
        \addplot[line width=0.25mm, blue, mark=*, mark size=1.5pt, error bars/.cd, y dir=both, y explicit,] coordinates {
                (1, 1) += (1,0.00) -=(1, 0.00)
(2,1.06959999999999) += (2,0.02) -=(2, 0.02)
(4,1.06319999999999) += (4,0.04) -=(4, 0.04)
(8,1.0722) += (8,0.04) -=(8, 0.04)
(16,1.07339999999999) += (16,0.03) -=(16, 0.03)
(32,1.0744) += (32,0.02) -=(32, 0.02)
(64,1.12539999999999) += (64,0.04) -=(64, 0.04)
(128,1.20499999999999) += (128,0.03) -=(128, 0.03)
(256,1.3282) += (256,0.02) -=(256, 0.02)
        };\label{plots:plot5}
        \addplot[smooth, line width=0.25mm,red, mark=square*, mark size=1.5pt, error bars/.cd, y dir=both, y explicit,] coordinates {
        (1, 1) += (1,0.00) -=(1, 0.00)
        (2,0.91) += (2,0.03) -=(2, 0.03)
        (4,0.91) += (4,0.05) -=(4, 0.05)
        (8,0.90) += (8,0.05) -=(8, 0.05)
        (16,0.90) += (16,0.04) -=(16, 0.04)
        (32,0.93) += (32,0.03) -=(32, 0.03)
        (64,0.92) += (64,0.05) -=(64, 0.05)
        (128,0.92) += (128,0.04) -=(128, 0.04)
        (256,0.96) += (256,0.03) -=(256, 0.03)
        };\label{plots:plot6}
        \draw[dashed] (1,1) -- (256, 1);

        \nextgroupplot[title=ResNet-50 (Tiny ImageNet), xlabel= Compression Ratio, ylabel = Speedup (GPU), xmin=1, ymax=1.25, xmax = 256, ymin=0.8, xtick={32,64,128,256}]

        \addplot[line width=0.25mm, blue, mark=*, mark size=1.5pt, error bars/.cd, y dir=both, y explicit,] coordinates {
        (1, 1) += (1,0.00) -=(1, 0.00)
(2,0.979399999999998) += (2,0.03) -=(2, 0.03)
(4,0.9266) += (4,0.01) -=(4, 0.01)
(8,0.9554) += (8,0.01) -=(8, 0.01)
(16,0.989799999999998) += (16,0.02) -=(16, 0.02)
(32,0.989599999999996) += (32,0.01) -=(32, 0.01)
(64,1.07499999999999) += (64,0.01) -=(64, 0.01)
(128,1.10239999999999) += (128,0.01) -=(128, 0.01)
(256,1.2222) += (256,0.01) -=(256, 0.01)
        };\label{plots:plot5}
        \addplot[smooth, line width=0.25mm,red, mark=square*, mark size=1.5pt, error bars/.cd, y dir=both, y explicit,] coordinates {
        (1, 1) += (1,0.00) -=(1, 0.00)
        (2,0.89) += (2,0.04) -=(2, 0.04)
        (4,0.89) += (4,0.02) -=(4, 0.02)
        (8,0.88) += (8,0.01) -=(8, 0.01)
        (16,0.89) += (16,0.03) -=(16, 0.03)
        (32,0.88) += (32,0.02) -=(32, 0.02)
        (64,0.91) += (64,0.02) -=(64, 0.02)
        (128,0.91) += (128,0.01) -=(128, 0.01)
        (256,0.92) += (256,0.01) -=(256, 0.01)
        };\label{plots:plot6}
        \draw[dashed] (1,1) -- (256, 1);

        \nextgroupplot[ylabel = Spatial Compression, xmin=1, xmax = 256, ymin=1, ymax=2.0,xtick={32,64,128,256}]
        \addplot[line width=0.25mm, blue, mark=*, mark size=1.5pt, error bars/.cd, y dir=both, y explicit,] coordinates {
        (1, 1)
        (2, 1.063)
        (4, 1.082)
        (8, 1.101)
        (16, 1.117)
        (32, 1.166)
        (64, 1.322)
        (128, 1.607)
        (256, 1.869)
        };\label{plots:plot5}
        \addplot[smooth, line width=0.25mm,red, mark=square*, mark size=1.5pt,] coordinates {
        (1, 1)
        (2, 1)
        (4, 1)
        (8, 1)
        (16, 1)
        (32, 1)
        (64, 1)
        (128, 1)
        (256, 1)
        };\label{plots:plot6}
        \draw[dashed] (1,1) -- (256, 1);
    \end{groupplot}
\path (current bounding box.north)--
      coordinate(legendpos)
      (current bounding box.north);
\matrix[
    matrix of nodes,
    anchor=south,
    draw,
    inner sep=0.1em,
    draw
  ]at([yshift=1ex]legendpos)
  {
    \ref{plots:plot5}& \DNNShifter (Structured Pruning) &[5pt]
    \ref{plots:plot6}& Unstructured Pruning\\};
\end{tikzpicture}
\caption{\bailey{Performance of \DNNShifter against other unstructured pruning methods as compression ratio increases; dashed line is baseline model performance. Each plot is the mean of five runs with confidence intervals of one standard deviation.}}
\label{new_exp2b}
\end{figure*}

Figure~\ref{new_exp2b} shows the CPU and GPU inference speed up and spatial compression achieved for increasing compression ratios. For \DNNShifterr, inference speed-up is defined as the inference latency of the baseline model over the pruned model for a given compression ratio. For the other unstructured pruning methods, we consider inference speed-up as the inference latency of the baseline model over the sparse model for a given compression ratio. Similarly, spatial compression is the in-memory size of the baseline model over that of the in-memory size of the pruned model (for \DNNShifterr) or of the sparse model (for unstructured pruning methods) for a given compression ratio. For both metrics, \DNNShifter provides improvements for all compression ratios, whereas unstructured pruning methods alone provide little or, in some cases, worse performance than the baseline model. 

\DNNShifter achieves up to 1.67x and 1.32x for VGG-16 and ResNet-50 on CPU and GPU, respectively, at no cost to the accuracy of the sparse model. 
The other unstructured pruning methods achieve a small speedup. However, the speedup varies due to irregular memory access in sparse models~\cite{sparse_perf}. \DNNShifter spatially prunes the sparse parameters and thus is not affected by structural irregularity. As the compression ratio increases, more sparse parameters are removed, resulting in a smaller model with lower CPU and GPU inference times.

\DNNShifter at a compression ratio of 256 achieves a spatial compression on the sparse model of 5.14x and 1.87x for VGG-16 and ResNet-50, respectively. ResNet-50 has a lower spatial compression ratio as DNNShifter only removes linear connections using structural pruning. As such, any skip connections or downsampling layers in ResNet-50 are not pruned as it will impact model accuracy~\cite{prunenet}. 

\begin{tcolorbox}[
    width=0.485\textwidth,
    enhanced,
    colframe=black!50!black,
    colback=gray!5,]
\bailey{
\textbf{Observation 3:}
\DNNShifter reduces inference latency and sparse model sizes in memory without losing accuracy. This is in contrast to sparse models obtained from unstructured pruning that are unsuitable for edge environments since they have poor inference performance and no spatial compression.}
\end{tcolorbox}

\subsubsection{Comparing accuracy against pruned models}
\label{subsec:new_exp3}
This study contrasts \DNNShifter against other structured pruning methods. Specifically, \DNNShifter is demonstrated to have comparable accuracy to the original model while producing similarly sized or smaller pruned models than other structured pruning methods. \DNNShifter is compared against RepVGG, EasiEdge, \bailey{ProsPr}, and two classic structured pruning methods: similarities-aware and $l^{1}$ norm. 

RepVGG, as described in Section~\ref{subsec:new_exp1}, creates VGG-like architectures by re-parameterising ResNet. In this study, RepVGG-A0 is the pruned VGG-16 model obtained from the baseline RepVGG-B0~\cite{repvgg}. EasiEdge is a recent structured pruning method that creates pruned models for edge deployment. \bailey{ProsPr is another modern pruning method that learns which weights to prune within the first few steps of optimisation~\cite{prospr}. In this study, we use the structured pruning variation that prunes channels. }
Pruning using $l^{1}$ norm is a classic structured pruning method that ranks the importance of each channel using $l^{1}$ norm and then prunes the lowest value channels~\cite{classical_struc}. Similarities-aware is another classic structured pruning method that removes channels with similar outputs~\cite{classical_struc}.

Table~\ref{tab:new_exp3} shows the accuracy change of the pruned model and the total parameter count after pruning the baseline VGG-16 on CIFAR-10 using different structured pruning methods. The table is organised in descending order of parameter count, where the baseline VGG-16 models are considered first and increasingly pruned models towards the bottom. EasiEdge \bailey{and ProsPr} models are denoted using a prune degree as a percentage. Prune degree is the percentage of pruned parameters from the baseline. For example, \bailey{EasiEdge-25\% prunes VGG-16 by 25\%}. \DNNShifter models are denoted using the compression ratio. For example, DNNShifter-2x has a compression ratio of 2, equivalent to a prune degree of 50\%.

\begin{table}[t]
\caption{
	Accuracy change and parameters of pruned VGG-16 models on CIFAR-10 for different structured pruning methods. \bailey{Values in bold are the best results for each category. Underlined values are second best for each category.}}
	\label{tab:new_exp3}
	\centering
\begin{tabular}{lrr}
\hline
 & \multicolumn{1}{r}{Acc. Change (\%)} & \multicolumn{1}{r}{Params. (M)} \\ \hline
VGG-16 (Baseline)                      & -                                                  & 14.72              
                                   \\
VGG-16 (Sim.-aware)                  & -5.82                                              & 11.04                                    \\ 
VGG-16 ($l^{1}$ norm)                     & -5.14                                              & 9.57                                     \\ 
RepVGG-A0                              & -0.40                                              & 8.30                                     \\ 
EasiEdge-25\%                          & +0.02                                              & 8.29                                     \\ 
$l^{1}$ norm + Sim.-aware & -5.30                                              & 8.10                                     \\ 
\DNNShifterr-2x                          & +0.18                                              & 7.54                                     \\ 
\bailey{ProsPr-80\%}                         & \bailey{+0.01}                                              & \bailey{4.01}                                     \\ 
\DNNShifterr-4x                          & +0.28                                              & 3.86                                     \\ 
EasiEdge-50\%                          & +0.13                                              & 3.68                                     \\ 
\bailey{ProsPr-90\%}                          & \bailey{+0.04}                                              & \bailey{2.00}                                     \\ 
\DNNShifterr-16x                         & \textbf{+0.40}                                     & 1.02                                     \\ 
\bailey{ProsPr-95\%}                         & \bailey{-0.28}                                     & \bailey{1.00}                                     \\ 
EasiEdge-80\%                          & -0.22                                              & 0.76                                     \\ 
EasiEdge-85\%                          & -0.51                                              & \underline{0.46}                                     \\ 
\DNNShifterr-64x                         & \underline{+0.33}                                              & \textbf{0.21}                            \\ \hline
\end{tabular}
\end{table}

Both classic structured pruning methods \bailey{showed more than 5\% accuracy reduction} with a pruning degree of 35\% or less. Combining the two methods allows for a similar accuracy loss but up to a pruning degree of 45\%. RepVGG-A0 achieves the same pruning degree as the combined classic methods while only dropping 0.4\% model accuracy. However, RepVGG does not have a smaller model variant than RepVGG-A0. \DNNShifter and EasiEdge produce smaller models with better accuracy than the baseline model. \DNNShifterr-16x has the best accuracy improvement with a 0.4\% gain, where a similarly sized EasiEdge-80\% lost 0.22\% accuracy. The smallest EasiEdge variant, namely EasiEdge-85\%, with 0.46M parameters, 0.51\% loss in accuracy, whereas \DNNShifterr-64x is over twice as small, with 0.21M parameters and gains 0.33\% accuracy. \bailey{ProsPr maintains a positive accuracy change up until models of size 2M parameters; however, accuracy remains lower than \DNNShifter at all model sizes.}

\begin{tcolorbox}[
    width=0.485\textwidth,
    enhanced,
    colframe=black!50!black,
    colback=gray!5,]
\bailey{
\textbf{Observation 4:}
\DNNShifter produces smaller and more accurate pruned models than other structured pruning methods.}
\end{tcolorbox}
\subsubsection{Pruning time against structured pruning methods}
\label{subsec:new_exp4}
\begin{figure}[!t]

\begin{tikzpicture}
\begin{axis}[  
    ybar,  
    enlargelimits=0.15,  
    ylabel={Pruning Time (s)},  
    xlabel={Structured Pruning Method},  
    height=5.3cm,  
    width=8.5cm,  
    ymode=log,  
    log origin=infty,  
    xtick=data, 
    x tick label style={rotate=45, anchor=north east},  
    symbolic x coords={DNNShifter, $l^{1}$ norm, EasiEdge, RepVGG, NTK-SAP$^{\dag}$},  
    extra y ticks=0.5,  
    extra y tick labels={},  
    axis background/.style={fill=gray!10},  
]  
    \addplot[ybar,fill=blue!40] coordinates {
        (DNNShifter,0.12) 
        ($l^{1}$ norm,3923) 
        (EasiEdge,4740) 
        (RepVGG,8) 
        (NTK-SAP$^{\dag}$, 544)
    };  
\end{axis}

\end{tikzpicture}
\caption{Average pruning time of a ResNet model using structured pruning.\\\bailey{$\dag$NTK-SAP unstructured pruning with \DNNShifter Phase 2 structured pruning.}}
\label{fig:new_exp4}
\end{figure}
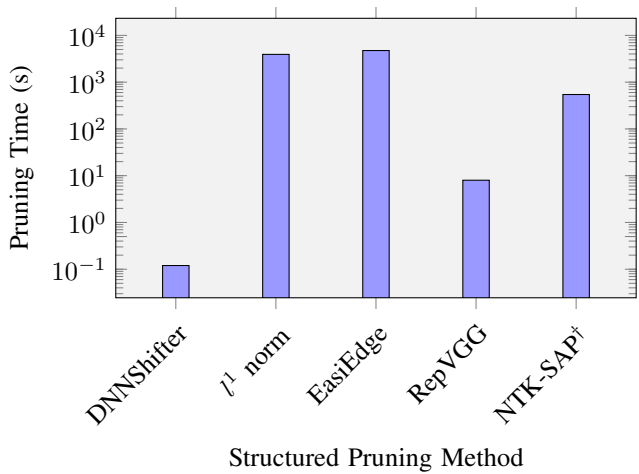

Figure~\ref{fig:new_exp4} shows the pruning time in seconds of various structured pruning methods to prune a ResNet model. 
$l^{1}$ norm requires almost 3,923 seconds to prune and fine-tune the model. 
EasiEdge does not require fine-tuning, but the ranking process it employs using Taylor approximations is exhaustive, thus requiring 4,740 seconds. 
RepVGG does not require ranking. Instead, it re-parameterises the model, which only requires 8 seconds. 
Although this is a relatively small cost, an 8-second overhead per training round may equate to a substantial overhead for certain use cases.
For example, consider pruning during the rounds of federated learning~\cite{prunefl}. \bailey{NTK-SAP requires 20 epochs of pre-training to generate an unstructured pruning mask resulting in 544 seconds of overhead.}
\DNNShifter prunes a model in a sub-second time frame. For ResNet, it is on an average of 120 ms\bailey{, or less than 3 frames for a 30 frames/second real-time edge video analytics application~\cite{realtime}, as opposed to tens of seconds to minutes of downtime with existing approaches}. 

\begin{tcolorbox}[
    width=0.485\textwidth,
    enhanced,
    colframe=black!50!black,
    colback=gray!5,]
    \bailey{
\textbf{Observation 5:}
\DNNShifter enables (near) real-time structured pruning of DNN models and is at least one order of magnitude faster than other structured pruning methods.}
\end{tcolorbox}

\subsection{Performance of model switching (Phase 3)}
\label{subsec:results_phase3}

Model switching enables an application to respond to changing runtime conditions by selecting a suitable model for inference from a pruned model portfolio. This study compares the in-memory compression and model switching method of \DNNShifter against the classic methods, namely model ensemble and Dynamic-OFA. The model ensemble method hosts simultaneous models in memory, and Dynamic-OFA uses a smaller sub-network within a single DNN to match operational demands.

Table~\ref{tab:new_exp5} compares runtime switching of \DNNShifter against model ensemble and Dynamic-OFA. \DNNShifter has a portfolio of four pruned VGG-16 models obtained in Section~\ref{subsec:new_exp1} with an accuracy range of 91.64-93.71\%\bailey{, model portfolio size of 30.4-66.1MB,} and a \bailey{CPU} inference speedup of 1.20-1.67x. A model ensemble of the same four VGG-16 models and a Dynamic-OFA model are also noted. Memory utilisation is the size of the model portfolio in memory. \DNNShifterr's memory utilisation is variable since inactive models are further compressed in memory (Section~\ref{sec:system}). However, in the model ensemble method, all models are uncompressed and hosted in memory. Similarly, Dynamic-OFA maintains the entire model in memory, even though only a smaller sub-network may be used during inference. \DNNShifter utilises as little as \bailey{3.8x} less memory for its model portfolio compared to the model ensemble method.

Decision overhead is the wall clock time for a model-switching method to select a model from the portfolio. For example, the model ensemble method runs inference on all models in the portfolio and then chooses the output with the highest confidence. On the other hand, Dynamic-OFA selects one DNN configuration to run inference and then re-configures the DNN to that selection. \DNNShifter inflates the appropriate model from in-memory and has an average decision overhead of 43 ms\bailey{, which is up to 11.9x faster than Dynamic-OFA.}

\begin{tcolorbox}[
    width=0.485\textwidth,
    enhanced,
    colframe=black!50!black,
    colback=gray!5,]
\bailey{
\textbf{Observation 6:}
\DNNShifter provides model switching with lower memory and decision overheads compared to existing methods. }
\end{tcolorbox}

\begin{table}[htp]
\caption{
	Memory utilisation and decision overhead of using model switching methods for VGG-16. \bailey{Relative values to \DNNShifter are recorded in parentheses.}}
	\label{tab:new_exp5}
	\centering
\begin{tabular}{lrr}
\hline
& \multicolumn{1}{r}{Mem. Util. (MB)} & \multicolumn{1}{r}{Overhead (ms)} \\ \hline

\footnotesize{Model Ensemble}~\cite{ensemble}  & 115.8 \bailey{(1.8-3.8x)}                                              & 49 \bailey{(1.14x)}                                           \\
\footnotesize{Dynamic-OFA}~\cite{dynamic-ofa}     & 56.2 \bailey{(0.9-1.8x)}                                              & 512 \bailey{(11.9x)}                                       \\ \hline
\textbf{\DNNShifterr}      & 30.4-66.1                                                & 43                                          \\ \hline
\end{tabular}
\end{table}


\section{Discussion and Conclusion}
\label{sec:conclusions}
Deploying production-quality DNNs in resource-constrained environments is essential for facilitating edge machine learning. Model compression offers techniques to derive model variants from a production-quality model suited for resource-constrained environments. However, \bailey{obtaining model variants that preserve accuracy and can be compressed to reduce the resource footprint and achieve low inference latencies is challenging.} Moreover, existing research has limited focus on adapting model variants to changing runtime conditions. 

\DNNShifter addresses the above concerns by developing an end-to-end framework that incorporates a novel pruning method and a time and resource-efficient pipeline for model training, compression, and runtime switching. \DNNShifter prepares model variants magnitudes faster than state-of-the-art neural architectural search, thus facilitating rapid and on-demand model deployments at the edge. The pruned model variants maintain the same accuracy as their production quality counterparts. They are suited for edge deployments since they are lightweight and adaptable to runtime conditions.

\bailey{\DNNShifter was designed to accommodate existing ML training and inference pipelines. \DNNShifter does not introduce any extra hyperparameters or dependencies other than requiring a user-specified maximum portfolio size. The structured pruning method of \DNNShifter can easily be used: (1) for one-time optimisation to pre-existing pre-trained DNN models, (2) in conjunction with the other phases to create a training and inference pipeline from scratch, or (3) in any combination of \DNNShifterr's phases. Thus, \DNNShifter is easily transferable to existing ML applications and products.}

\bailey{\DNNShifter is primarily limited by the high computation cost of training sparse models. There is potential for structured pruning to be conducted at the initialisation of the model (before training) with minimal accuracy loss~\cite{pat, cai2022structured}. This will be explored in the future.}

\section*{Acknowledgements}
This research is funded by Rakuten Mobile, Japan.

\bibliographystyle{IEEEtran} 
\bibliography{main}

\end{document}